\documentclass[conference,a4paper]{IEEEtran}
\IEEEoverridecommandlockouts
\usepackage{cite}
\usepackage{amsmath,amssymb,amsfonts}
\usepackage{algorithmic}
\usepackage{graphicx}
\usepackage{textcomp}
\usepackage{xcolor}
\usepackage[hyphens]{url}
\usepackage{comment}
\usepackage[english]{babel}
\usepackage[T1]{fontenc}
\usepackage{lmodern}
\usepackage{newtxtext,newtxmath}
\usepackage{amsmath}
\usepackage{graphicx}
\usepackage[caption=false]{subfig}
\usepackage{makeidx}
\usepackage{multicol}
\usepackage{booktabs}
\usepackage{multirow}
\usepackage{enumitem}
\usepackage[numbers]{natbib}
\usepackage{changepage}

\def\BibTeX{{\rm B\kern-.05em{\sc i\kern-.025em b}\kern-.08em
    T\kern-.1667em\lower.7ex\hbox{E}\kern-.125emX}}
\begin{document}
\title{Multichannel Attention Networks with Ensembled Transfer Learning to Recognize Bangla Handwritten Charecter\\}

\author{
    \IEEEauthorblockN{ Farhanul Haque, Md. Al-Hasan, \\ Sumaiya Tabssum Mou}
    \IEEEauthorblockA{
        \textit{Department of Computer Science and Engineering} \\
        \textit{Bangladesh Army University of Science and Technology (BAUST)}, \\Saidpur Cantonment, Bangladesh \\
        % Email if needed: farhanul@example.com
    }
    \and
    \IEEEauthorblockN{Abu Saleh Musa Miah*, Jungpil Shin*}
    \IEEEauthorblockA{
        \textit{Department of Computer Science and Engineering} \\
        \textit{School of Computer Science and Engineering} \\
        \textit{University of Aizu}, \\Aizuwakamatsu, Fukushima, Japan \\
        Email: abusalehcse.ru@gmail.com,jpshin@u-aizu.ac.jp
    }
    \and
    \IEEEauthorblockN{Md Abdur Rahim}
    \IEEEauthorblockA{
        \textit{Department of Computer Science and Engineering} \\
        \textit{Pabna University of Science and Technology}, Rajapur, Pabna, Bangladesh \\
        % Email if needed: rahim@example.com
    }
   }

\maketitle
\begin{abstract}
The Bengali language is the 5th most spoken native and 7th most spoken language in the world, and Bengali handwritten character recognition has attracted researchers for decades. However, other languages such as English, Arabic, Turkey, and Chinese character recognition have contributed significantly to developing handwriting recognition systems. Still, little research has been done on Bengali character recognition because of the similarity of the character, curvature and other complexities. However, many researchers have used traditional machine learning and deep learning models to conduct Bengali hand-written recognition. The study employed a convolutional neural network (CNN) with ensemble transfer learning and a multichannel attention network. We generated the feature from the two branches of the CNN, including Inception Net and ResNet and then produced an ensemble feature fusion by concatenating them. After that, we applied the attention module to produce the contextual information from the ensemble features. Finally, we applied a classification module to refine the features and classification. We evaluated the proposed model using the CAMTERdb 3.1.2 data set and achieved 92\%  accuracy for the raw dataset and 98.00\%  for the preprocessed dataset. We believe that our contribution to the Bengali handwritten character recognition domain will be considered a great development. 
\end{abstract}
\section{Introduction}
Automatic Handwritten Character Recognition has become a favourite research do-main to researchers in recent years for numerous applications, such as Optical character recognition (OCR). OCR is a formula to convert the picture of handwritten or picture of printed text into machine-encoded text \cite{sayeed2021bengalinet}. Many researchers proposed artificial intelligence techniques \cite{ kibria2020creation} for recognizing various text in historical documents, printed documents, number plates of vehicles, business cards, passports, bank checks, handwritten characters etc. The main challenges of handwritten character recognition are the numerous writing fashion for a specific character and the different types of handwriting styles by various writers. In addition, there are a lot of characters which have similar patterns. Some of the complicated handwriting scripts recognize various ways of writing words. Handwritten character r recognition compares more intriguingly with typed character types.
Moreover, handwritten characters written by various writers are not identical but vary in characteristics such as size and shape. Numerous variations in the writing manner of the same characters make the recognition task more challenging. The overlaps and the interconnections of the nearest characters further complicate the task. The substantial variety of writing styles, writers, and the complicated features of the handwritten characters are very challenging for perfectly recognizing the handwritten characters.
\par
Also, the character is not similar in the whole world, and according to a research report in 2020, there are around 7117 languages, and there have different types of character with different and similar style and fashions \cite{eberhard2015ethnologue}.  Researchers have been working to recognize handwritten characters from different specific languages with specific letters or alphabets. Because of the language-specific interest of research, all languages did not get equivalence development for this case. Though some languages got huge contributions from many researchers, many languages did not make any significant contribution to recognising handwritten text or characters, and the Bengali language is one of them.  The Bengali language is an Indo-Aryan language. It is the fifth most spoken native language and seventh most spoken language in the world, and around 228 used this language as their first language. In addition, approximately 37 million people worldwide use this language as their second language for speaking \cite{miah2022bensignnet}. Bengali handwritten character recognition is one of the most complex tasks because the number of characters in this language is not the same as the other language literature.  The dataset for the handwritten character recognition of the Bangla language included 50 basic character’s but this number is not fixed like in another well-known language because of the compound characters.  By including the compound characters of this language, it takes approximately 400 handwritten character’s which yielded so much complex work in the world \cite{das2014benchmark} using the conventional feature and machine learning approach \cite{miah2020motor, miah2022natural}. However, deep learning-based models such as CNN \cite{miah2023rotation, rahim2020hand} transfer learning, attention-based model \cite{miah2023multistage, miah2023dynamic} and transformer \cite{shin2023korean} can be taken care of in more than 1000 classes datasets with complex neural networks. Still many difficulties in dealing with Bangla handwritten characters due to the complexity of cursive in shape, and especially compound characters, which make it complicated. Also, many researchers proposed different algorithms to recognize Bengali handwritten characters. Still, their performance accuracy and system efficiency are not satisfiable \cite{hoq2019bangla, dinajpur2019handwritten, begum2017recognition, rahman2015bangla, das2010handwritten, rabby2019ekush, khan2014handwritten,miah2023multi,kabir2023investigating,zobaed2020real,joy2020multiclass} In addition, attention-based deep learning models are numerously used in various computer vision fields. Still, few are employed in Bengali handwritten recognition \cite{miah2023multistage, miah2023dynamic}. 
\par
To solve the efficiency and lower performance accuracy problem, we proposed an ensemble convolutional neural architecture through multi-channel attention in the study.  In the processing, first, we take input of our images from the dataset, then we dived the dataset into a training and testing set. Then each of the sets, we employed a preprocessing model for augmentation and colour processing. Then we used the Ensemble module, where we included GoogLeNet and ResNet architecture to extract different features with two branches from the handwritten character dataset.  The preprocessed image was fed separately to the two pre-trained models, which discovered and investigated the important feature from that. Then after concatenating the two-branch feature and producing the ensemble feature, we used a multichannel attention approach we employed here, which was used in the previous study \cite{miah2022bensignnet, rahim2020hand}. This attention block sets the efficient and adaptable weight to the ensemble feature maps for choosing the more strong and potential features which can be fed into the classification module to improve performance accuracy. The main idea of the attention approach is to investigate the global context information of the images because of the diverse significance of the properties, and it can extract the class-specific important information. After that this attention-oriented new feature, finally, a classification module is used to refine the feature and classification purposes. The experimental evaluation demonstrated the proposed model boosted the performance accuracy by a significant margin.

\section{Related Work}
Machine Learning and Deep Learning algorithm proved their excellency in various fields such as EEG-based various disease and event classification \cite{miah2017motor,miah2019eeg,miah2019motor,miah2022natural,joy2020multiclass,zobaed2020real,miah2021event,miah2022movie,kabir2023investigating}, object detection \cite{kibria2020creation}, vision-based gesture and human activity classification \cite{miah2023multistage, miah2023dynamic, shin2023korean,hossain2023stochastic} and disease detection \cite{miah2021alzheimer,kafi2022lite}.
Based on the excellency many researchers have been working to develop a Handwriting recognition system using various statistical features incorporating machine learning and deep learning method \cite{miah2023multistage, miah2023dynamic, shin2023korean, rahim2020hand,  hoq2019bangla, dinajpur2019handwritten, rabby2019ekush, khan2014handwritten}. 
In recognition of the Bangla character, many previous works have been developed for the Bangla digit, including ten digits. A few works are also available for Bangla handwritten character recognition, which contains 50 characters. Moshiur Rahman et el. 's "Handwritten Bengali Character Recognition Through Geometry Based Feature Extraction" \cite{hoq2019bangla} their proposed method achieved an average recognition rate of 84.56\% using SVM and 74.47\% using ANN. "Bangla Hand-written Character Recognition: an overview of the state-of-the-art classification Algorithm with new dataset" \cite{hoq2019bangla} created by Md Nazmul Hoq et. el they have shown that the ANN has achieved 93.8\% and Logistic Regression has achieved (86.8\%) that outperform all other classification algorithms. Halima Begum et al., "Recognition of Handwritten Bangla Characters using Gabor Filter and Artificial Neural Network" \cite{begum2017recognition} experimented with their own dataset that was the collaboration of 95 volunteers and their proposed technique achieved 68.9\% without feature extraction and 79:4\% with feature extraction. \par "Bangla Handwritten Char-acter Recognition using Convolutional Neural Network" \cite{rahman2015bangla} achieved test accuracy of 85.36\% using their own dataset. In "Handwritten Bangla Basic and Com-pound character recognition using MLP and SVM classifier" \cite{das2010handwritten}, handwritten Bangla character recognition with MLP and SVM has been proposed, and they achieved around 79.73\% and 80.9\% of recognition rate, respectively. AKM Sha-harder et el. "Ekush: A Multipurpose and Multitype Comprehensive Database for Online Off-Line Bangla Handwritten Characters" \cite{rabby2019ekush} achieved 97.73\% for the Ekush dataset. A sparse representation classifier is applied for Bangla digit recognition in \cite{rabby2019ekush}, where 94\% of accuracy was achieved for handwritten digit recognition.

\section{Dataset}
For our purpose, we used CAMTERdb 3.1.2. This dataset has 12000 images of 50 classes. Each class contain 240 BMP format 3-channel image. Most of the images have noise-free pixels, almost correct labelling, and no overwriting characters images. 
\begin{figure}[htb]
    \centering
\includegraphics[width=0.5\linewidth]{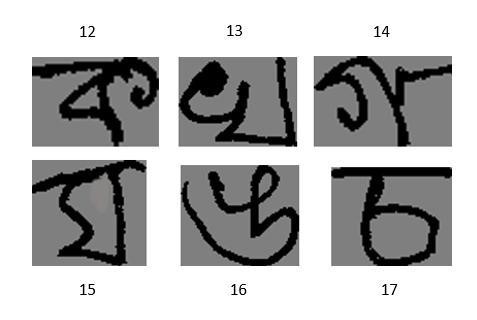}
    \caption{Images with respective class numbers.}
    \label{img_class_num}
\end{figure}
No matter which state in the dataset, we have to preprocess it to make it easier for the architecture to learn accurately \cite{sayeed2021bengalinet,das2014benchmark,das2015handwritten}. Sample images of the usage dataset are shown in Figure \ref{img_class_num}

\section{Proposed Model}
\begin{figure}[htb]
    \centering
    \includegraphics[width=1\linewidth]{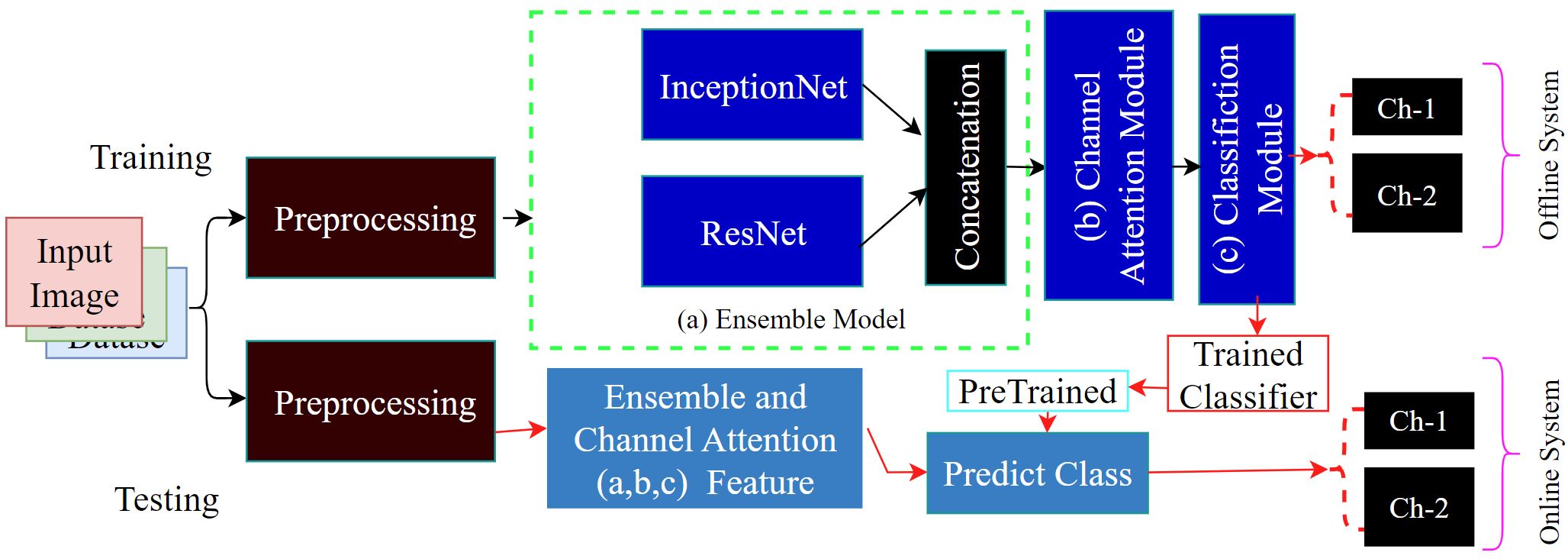}
    
    \label{main_figure}
    \caption{Working flow architecture}
\end{figure}

In this study, we proposed an ensemble convolutional neural architecture to recognise Bengali handwritten characters through multi-channel attention. Figure \ref{main_figure} demonstrated the working flow graph of the proposed model. In the flowchart given that first, we take input of our images from the dataset, then we dived the dataset into a training and testing set. Then each of the sets we employed a pre-processing model where we inverted the color of every image to reduce the computation cost. Then those images were converted into grayscale format. The dataset was augmented by random rotation to fix black corners. Then we used the Ensemble module, which included GoogLeNet and ResNet architecture to extract different features with two branches from the handwritten character dataset.  The preprocessed image was fed separately to the two pre-trained models, which discovered and investigated the important feature from that. Then after concatenating the two-branch feature and producing the ensemble feature, a multichannel attention approach we employed here, which is used in the previous study~\cite{miah2022bensignnet,rahim2020hand}.

 This attention block sets the efficient and adaptable weight to the ensemble feature maps for choosing the more strong and potential features which can be fed into the classification module to improve performance accuracy. The main idea of the attention approach is to investigate the global context information of the images because of the diverse significance of the properties, and it can extract the class-specific important information. After this attention-oriented new feature, finally, a classification module is used to refine the feature and classification purposes. The experimental evaluation demonstrated that the proposed model boosted the performance accuracy significantly. The workflow of the proposed model is demonstrated in Figure \ref{sam_output_pre}. 

\subsection{Preprocessing Module}
\begin{figure}[htb]
    \centering
    \includegraphics[width=1\linewidth]{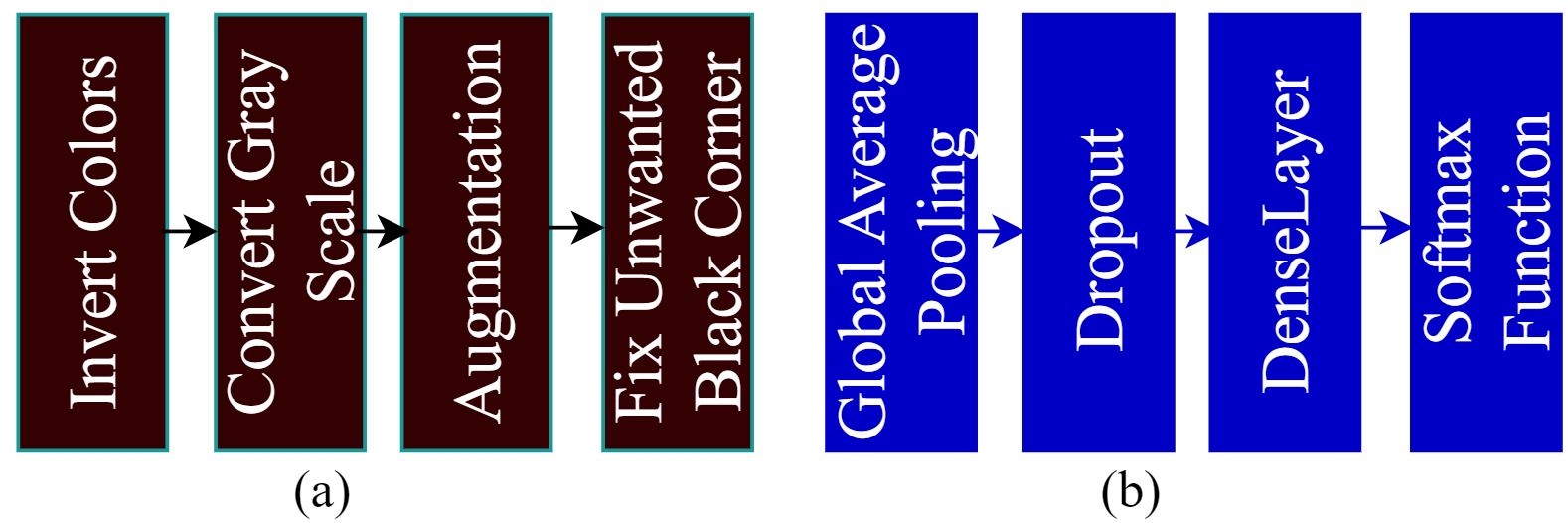}
    \caption{Preprocessing and classification module architecture}
    \label{pre_class_module}
\end{figure}

Before extracting deep learning-based features, we employed a preprocessing module in this study. Figure \ref{pre_class_module}(a) demonstrated the preprocessing module where first applied the inverted colours then applied to a converter to convert the RGB image into a grayscale image because the grey background contains only black and white, which reduces the computational complexity of the system.

\begin{figure}[htb]
    \centering
    \includegraphics[width=0.5\linewidth]{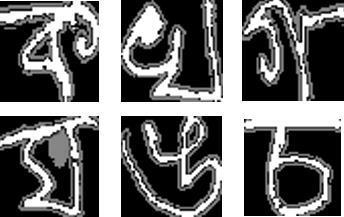}
    \caption{Sample output after preprocessing}
    \label{sam_output_pre}
\end{figure}

After that, we applied the augmentation technique for the rotation of the image into the different orientations, and finally, we fixed the unwanted black corner of the images. Figure \ref{sam_output_pre} visualized the sample output of the preprocessing images. 

\subsection{Ensemble Module}
The study proposes an ensembled convolutional neural network with a channel attention model to classify the handwritten character in the Bengali language. In designing the architecture, two pre-trained CNN models we introduce here including GoogleNet and ResNet transfer learning~\cite{szegedy2015going,simonyan2014very,jian2016deep,larsson2016fractalnet,he2016deep}. We extract-ed separately from these two models with two branches and then concatenated them to produce the ensemble features described below. 
\\

\subsubsection{GoogleNet Architecture}

Machine learning has offered a wide range of techniques and Application-based models and huge data sizes for processing and problem-solving. Deep learning, a subset of Machine Learning, is usually more complex. So, thousands of images are needed to get accurate results \cite{szegedy2015going}.\\
The GoogleNet architecture was proposed by researchers from Google in 2014. The research paper was titled "Going Deeper with Convolutions". This architecture was a winner at the ILSVRC 2014 classification of image challenge. It has shown a significant decrease in error rate if we compare it with to previous winner AlexNet. GoogleNet architecture uses techniques such as 1x1 convolutions in the middle of the architecture and also global average pooling. his architecture is different. Using 1x1 convolution as intermediate and global average pooling creates a deeper architecture. This architecture is 22 layers deep. Google Net uses 1x1 convolutions in architecture. The convolutions are used to decrease the architecture's parameters (weights and biases). For a convolution when the filter number is 48 but the filter size is 5 x 5, the number of computations can be written as (14 x 14 x 48) x (5 x 5 x 480) = 112.9M. On the other side, for the 1x1 convolution with 16 filters a number of computations can be written as.  (14 x 14 x 16) x (1 x 1 x 480) + (14 x 14 x 48) x (5 x 5 x 16) = 5.3M. The inception architecture in GoogleNet has some intermediate classifier branches in the middle of the architecture. These branches are used only while training. These branches consist of a 5x5 average pooling layer with a stride of 3, one 1x1 convolutions with 128 filters, two fully connected layers of 1024 outputs and 1000 outputs and a SoftMax classification layer.  
The generated loss of these layers throughout training added to the total loss with a weight of 0.3. The purpose of these layers is they help in combating the gradient vanishing problem and also these layers provide regularization. The architectural details of the auxiliary classifier can be described in the following steps (a) A 1x1 convolution with 128 filters for dimension reduction and ReLU activation. (b) An average polling layer of filter size 55 and stride is 3 value. (c) Dropout regularisation with dropout ratio = 0.7. (d) A fully connected layer with 1025 outputs and ReLU activation
\\
\subsubsection{ResNet Architecture}

ResNet comes from Residual Network which is one of the most efficient neural network architectures and is developed by researchers Aiming and his team \cite{he2016deep}. We are focusing on the network because it has the ability to train the architecture and achieve good performance in the computer vision-related research domain, such as object detection, segmentation, and classification. The objective is to develop of this system is to achieve efficiency to vanish the gradients using a deep neural network with minimum cost and time which can improve efficiency and performance.  The main concept of the idea was to implement a technique to pass the information from one layer to another by flowing directly without fetching intermediate deep layers. To do this they used a shortcut connection to pass the information between any two layers. As a consequence, this idea helps the gradients to flow easily within the network yielding mitigate the gradient vanishing problems. We can write the working procedure of this transfer learning using some steps such as (a) to extract feature it passes the input data through a series of sequential convolutional layers. (b) The extracted feature is then passed through the multiple residual connections which consist of the multiple convolutional layer and a skip connection. (c) the skip connection is able to pass the feature to some of the CNN layers or send it to the output block. (d) The final skip connection passed the data through a global average pooling layer which averages the data and aims to convert the matrix feature into vector form which is known as a pooled feature. (e) The last pooled feature fed the data into a fully connected layer to produce the final feature which is used here as a concatenation vector.  Bypassing the information through various deep layers allows for the training of a much deeper network compared to the previously existing systems yielding better efficiency and performance. 

\subsection{Channel Attention}
In the previous section, we discussed the two pre-trained models which we employed for achieving ensemble features which are concatenated and produced the final features. The pre-trained model also has strong pulling properties which vary from sample to sample dataset based on the several filters. Some research proved that sequentially adding the attention mechanism after the pre-trained model has a significant impact on the classification~\cite{hirooka2022ensembled,yan2017multibranch}.

\begin{figure}[htb]
    \centering
    \includegraphics[width=0.85\linewidth]{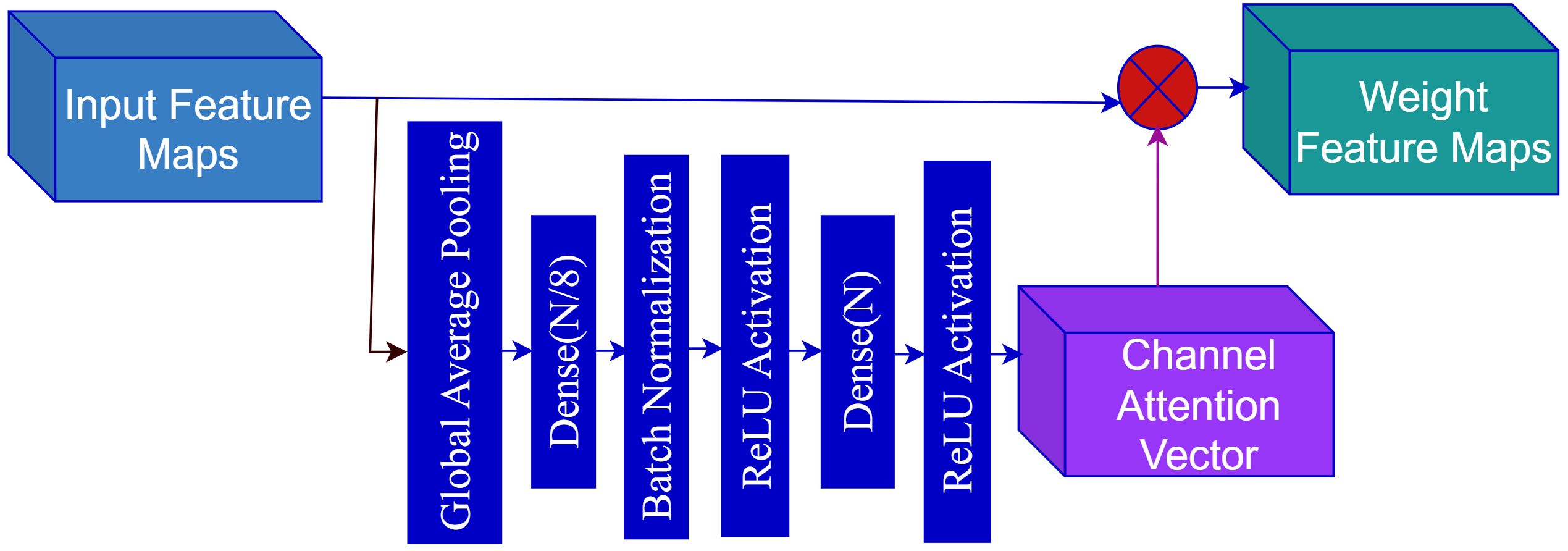}
    \caption{Channel attention of the proposed model}
    \label{fig:chanel_att_model}
\end{figure}
In our study, we fed the concatenated feature into the channel attention which runs through the global average pooling and then produces the channel-wise feature amount. After that, each channel is coupled with two convolution layers with the employed ReLU activation approach to achieve a value considered positive or 0. The main emphasis of the output value is that the positive value indicates the significant or potential information and 0 indicates the unnoticeable feature or less significant channels. We also added two extra ReLu function which helps to reduce the time and complexity to vanish the gradient. Figure \ref{fig:chanel_att_model} demonstrated the attention model which we utilized in the study where firstly we employed a global average pooling on the input feature which contained N channel.  Then we employed a dense layer which has N/8 channels and fed into the batch normalization layer which can able to solve the covariate shift problems. It also contributes to preventing the gradient vanishing challenges from becoming too small. After that, we used the ReLU action function and fed that into another Dense Layer where the N layer was added through the reusing ReLU activation function. We are focusing on the ReLU because its computational complexity is less than the other activation like sigmoid functions.  Once the neural architecture is too large or has many neurons then the significance of this activation function is easily understandable which can significantly reduce the time for training and testing times. In addition, it can also help us to faster converge into the fitting time.

\subsection{Classification Module}
In the classification module, we employed different deep layers to refine the attention based feature and the classification. Figure \ref{pre_class_module}(b) demonstrated the proposed classification module. We employed a global average pooling layer and then fed it into the dropout layer. After that, we applied a fully connected layer. Those fully connected layers contain the majority of parameters of many architectures that causes an increase in computation cost. But in our architecture, we first applied a method called global average pooling. This layer takes a feature map and averages it to 1x1. This also decreases the number of trainable parameters to 0 and improves the top-1 accuracy y 0.6%.
\\
\section{Performance Evaluation}
We evaluated the proposed model using a Bengali handwritten character dataset for investigating the superiority and effectiveness of the proposed model. In the below section firstly, we described the environmental setting of the project the performance accuracy. 
\\
\subsection{Environment Setting}
Our experimental environment was configured within Google Colaboratory. Cloud-based Collaboratory was a free Jupyter Notebook environment which required no extra setup or installs Python libraries. Each colab session was equipped with a virtual machine which provided 13GB of RAM either on a GPU, TPU or CPU processor. We have executed our experiment in a GPU processor. 
\\
\subsection{Experimental Results}
We observed our experiment in three parts. Table \ref{sample-table} demonstrated the performance of the proposed model with various experiments. We firstly experimented with the model with only GoogleNet which achieved 85.32\% accuracy, in the same way with ResNet it achieved 87.32\% accuracy. After that once we employed the ensemble including the attention and classification module it produced 98.00\% accuracy. 

\vspace{-5mm}
\begin{table}[htb!]
\centering % Center the table
\caption{Performance accuracy of the proposed model.}
\begin{tabular}{l|l|l} 
\toprule
Dataset & Model Name & Accuracy [\%] \\ \midrule
CAMTERdb 3.1.2 & Proposed model with GoogleNet & 85.32 \\
CAMTERdb 3.1.2 & Proposed model with ResNet & 87.32 \\
CAMTERdb 3.1.2 & Proposed Ensemble Model & 98.00 \\ 
\bottomrule
\end{tabular}
\label{sample-table}
\end{table}

\subsection{State of the Art Comparison}
Table \ref{comparison} demonstrated the state-of-the-art comparison of the proposed model. The table showed that our proposed model achieved 98.00\% accuracy, which is the best performance compared to all state-of-the art study those are mentioned here. In the previous study Halima Begum et al extracted features using a Gabor filter from the handwriting image then they applied an artificial neural network for the classification where they achieved 79.40\% accuracy with it ~\cite{begum2017recognition}. Das et al. extracted geometric features from the Bangla handwriting image and achieved 76.86\% accuracy with a machine learning algorithm\cite{das2014recognition}. Rahman et al. applied the CNN method for the feature extraction and the classification for the Bangla handwriting recognition and achieved 85.36\% accuracy \cite{rahman2015bangla}. Islam et al. extracted effective features using the modified syntactic model and achieved 95.00\% accuracy \cite{islam2005bengali}. The high-performance accuracy of the proposed model proved its superiority compared to the existing Bangla hand writing recognition system. 

\vspace{-5mm}
\begin{table}[!htb]
\centering % Center the table
\caption{Previous work with different methods} % Consistent with "methods" being plural
\begin{tabular}{l|l|l} 
\toprule
Author Name & Method  & Accuracy [\%] \\ \midrule
Halima Begum et al.~\cite{begum2017recognition} & Gabor Filter and ANN & 79.40\% \\
Nibaran Das et al.~\cite{das2014recognition} & Convex Hull Basic Feature (CHBF) & 76.86\% \\
Rahman et al.~\cite{rahman2015bangla} & Convolutional Neural Network (CNN) & 85.36\% \\ 
Islam et al.~\cite{islam2005bengali} & Modified syntactic method & 95.00\% \\ 
Proposed Model & MANETL & 98.00\% \\ 
\bottomrule
\end{tabular}
\label{comparison}
\vspace{-4mm}
\end{table}

\section{Conclusion}
We have recognized Bangla's handwritten characters using Multichannel Attention and Ensemble feature fusion architecture. In the processing we first applied a preprocessing model then we applied a two-pretrained model to extract separate features which we concatenated to produce the ensemble feature. Then we applied a channel attention model to highlight the effective feature and finally, we applied a classification module to classify and make a pre-trained model. Performance accuracy reported in the accuracy table proved the efficiency and effectiveness of the proposed model. For our future plan, we'll implement this to get real-time traffic camera recognition. We can also extend our work to the Bangla digit and Bangla joint alphabet. Finally, our work can be extended to recognise Bangla handwritten text such as handwritten letters, applications, exam scripts, and different forms also for a complete OCR system.
\let\cleardoublepage\clearpage
%\begin{thebibliography}{99}
%\bibliographystyle{plain}
%\bibliographystyle{unsrtnat}
\bibliography{main}
\bibliographystyle{plain}
\end{document}